%% file: main.tex
\documentclass[sigconf]{acmart}

\AtBeginDocument{%
  \providecommand\BibTeX{{%
    \normalfont B\kern-0.5em{\scshape i\kern-0.25em b}\kern-0.8em\TeX}}}

\setcopyright{acmcopyright}
\copyrightyear{2018}
\acmYear{2018}
\acmDOI{XXXXXXX.XXXXXXX}

\acmConference[Conference acronym 'XX]{Make sure to enter the correct
  conference title from your rights confirmation emai}{June 03--05,
  2018}{Woodstock, NY}
\acmPrice{15.00}
\acmISBN{978-1-4503-XXXX-X/18/06}

\input{packages.tex}

\input{command.tex}
\begin{document}

\title{Web News Timeline Generation with Extended Task Prompting}

\author{Sha Wang\(^{1}\), Yuchen Li\(^{1}\), Hanhua Xiao\(^1\), Lambert Deng\(^2\) Yanfei Dong\(^3\)
}
\affiliation{
\(^1\) Singapore Management University  \(^2\) DBS Bank Limited 
\(^3\) PayPal\\
\country{Singapore}
\fontsize{10}{10}\texttt{sha.wang.2021@phdcs.smu.edu.sg, yuchenli@smu.edu.sg, hhxiao.2020@phdcs.smu.edu.sg, lambertdeng@dbs.com, dyanfei@paypal.com}
}


\begin{abstract}
\import{./sections/}{abstract}
\end{abstract}



\begin{teaserfigure}
\vspace{-5mm}
  \includegraphics[width=\textwidth]{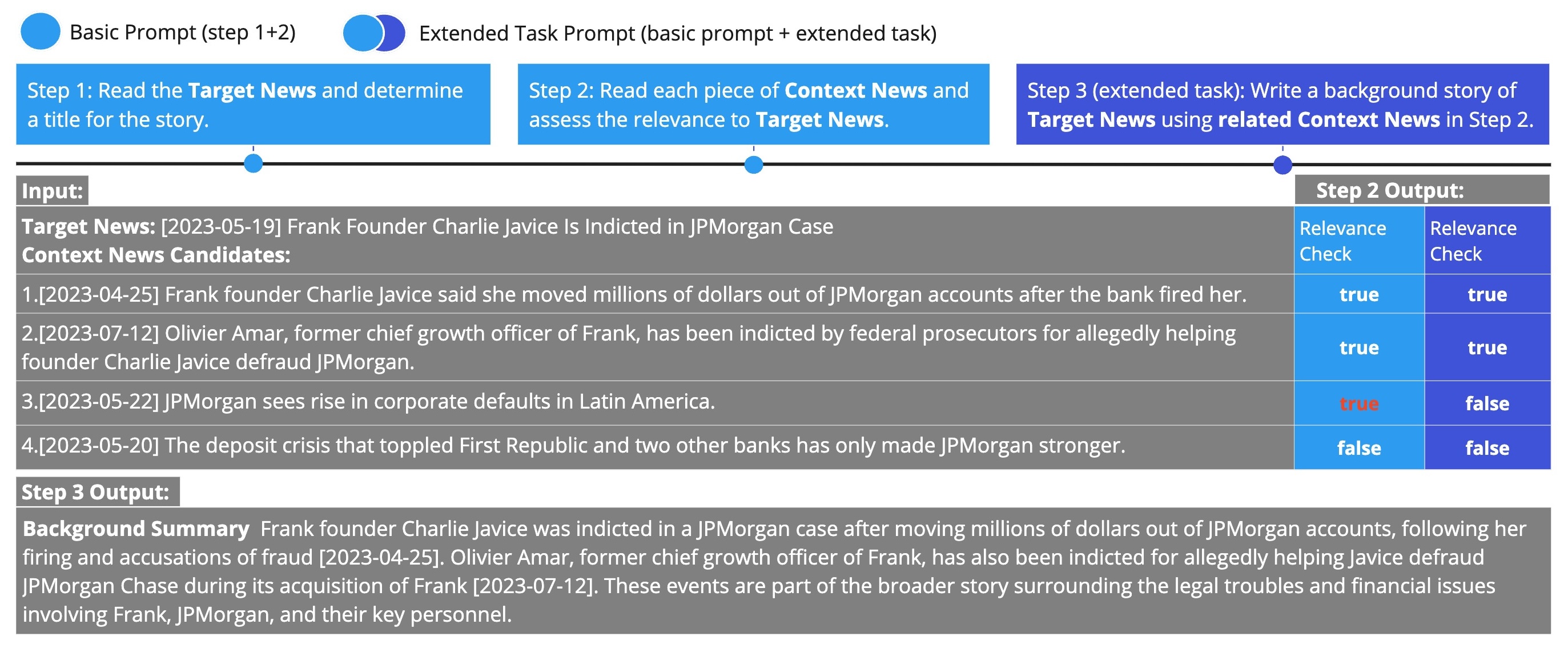}
  \caption{In this example, the target task is to assess the relevance of Context News by prompting a Large Language Model. To improve the performance of the target task, we incorporate an extended task in the prompt which writes a background story for Target News using relevant Context News identified in the target task.}
  \label{fig:teaser}
\end{teaserfigure}


\maketitle

\section{Introduction}
\import{./sections/}{introduction}
\section{Related Work}
\import{./sections/}{related_work}

\section{Architecture and Prompt Engineering}
\import{./sections/}{architecture}
\pagebreak
\section{Evaluation}
\import{./sections/}{evaluation}
\vspace{-3mm}
\section{Conclusion}
\import{./sections/}{conclusion}

\pagebreak
\bibliographystyle{ACM-Reference-Format}
\bibliography{main}

\end{document}

%% file: packages.tex
\usepackage{import}
\usepackage{amsmath}
\usepackage{comment}
\usepackage{xspace}

\usepackage{enumitem}
\usepackage{graphicx}
\usepackage[format=default,font=large,labelfont=bf]{caption}
\usepackage{changepage}
\usepackage{adjustbox}
\usepackage{listings}
\definecolor{codegreen}{rgb}{0,0.6,0}
\definecolor{codegray}{rgb}{0.5,0.5,0.5}
\definecolor{codepurple}{rgb}{0.58,0,0.82}
\definecolor{backcolour}{rgb}{0.95,0.95,0.92}

\lstdefinestyle{mystyle}{
    backgroundcolor=\color{backcolour}, 
    basicstyle=\ttfamily\footnotesize,
    breakatwhitespace=false,         
    breaklines=true,                 
    captionpos=t,                    
    keepspaces=false,                 
    numbersep=1pt,                  
    showspaces=false,                
    showstringspaces=false,
    showtabs=true,                  
    tabsize=1
}

%% file: command.tex
\makeatletter
\newcommand*{\rom}[1]{\expandafter\@slowromancap\romannumeral #1@}
\makeatother

\newcommand{\demosite}{\url{https://storyline.tembusu.link}}
\newcommand{\pluginsite}{\url{https://chrome.google.com/webstore/detail/news-storyline/hicnhclaebdfajlinkaijciloachmfgk/}}

\newcommand{\quoted}[1]{``\textit{#1}''}

\newcommand{\basicprompt}{\textit{basic prompt}\xspace}
\newcommand{\extendedprompt}{\textit{extended task prompt}\xspace}

%% file: sections/abstract.tex
The creation of news timeline is essential for a comprehensive and contextual understanding of events as they unfold over time. This approach aids in discerning patterns and trends that might be obscured when news is viewed in isolation. By organizing news in a chronological sequence, it becomes easier to track the development of stories, understand the interrelation of events, and grasp the broader implications of news items. This is particularly helpful in sectors like finance and insurance, where timely understanding of the event development—ranging from extreme weather to political upheavals and health crises—is indispensable for effective risk management. 
While traditional natural language processing (NLP) techniques have had some success, they often fail to capture the news with nuanced relevance that are readily apparent to domain experts, hindering broader industry integration.
The advance of Large Language Models (LLMs) offers a renewed opportunity to tackle this challenge. However, direct prompting LLMs for this task is often ineffective.
Our study investigates the application of an extended task prompting technique to assess past news relevance. We demonstrate that enhancing conventional prompts with additional tasks boosts their effectiveness on various news dataset, rendering news timeline generation practical for professional use. This work has been deployed as a publicly accessible browser extension which is adopted within our network.

%% file: sections/introduction.tex
In the realm of financial risk management, noteworthy events frequently occur within a condensed time period and impact numerous stakeholders, posing challenges in effectively monitoring their progression. A case in point is the failure of Silicon Valley Bank (SVB), where the bank experienced depositor distress, rapidly escalated to a bank run, and culminated in an FDIC takeover, all within a single weekend. 
In the realm of financial risk management, 
Conversely, the true significance of certain events may only become apparent when evaluated in an extended time frame. For instance, the collapse of SVB has been attributed by many to the executive leadership's inadequate grasp of the balance sheet impact of high interest rates, an unseen circumstance in recent decades. Had the executives seen historical news of how banks adjusted their portfolio in 1980s, the bank's failure may well have been avoided. This underlines the importance of monitoring event progression by considering historical news data, to achieve a more comprehensive and accurate understanding of the event's developments. News timeline generation exemplifies efforts in this direction.

However, the endeavor to identify relevant past news for constructing a coherent timeline presents considerable challenges. In an era of information explosion, locating the appropriate news articles is akin to searching for a needle in a haystack. Additionally, numerous subtleties dictate the relevance of news, subtleties often discernible only to domain experts. Such complexities render traditional natural language processing (NLP) methods frequently inadequate for this task.

Recent advancements in Large Language Models (LLMs)~\cite{gpt-3.5-turbo,openai2023gpt4,touvron2023llama} have spurred a reevaluation of our approach to comprehending and summarizing the development of events. 
Extensive pre-training of LLMs enables them to detect nuanced subtleties that used to require domain expertise. In this research, we explore the efficacy of prompt engineering techniques for language models, particularly focusing on the task of timeline generation from a series of news reports. The primary input includes a target news report, alongside a compilation of context news candidates. 
The model's task is to determine the relevance of each candidate news with respect to the target news. However, we find that the \basicprompt method to determine relevance of candidate news (the target task) often yields unsatisfactory results as illustrated in Figure~\ref{fig:teaser}. In our study, we present \extendedprompt, where an extended task of summarizing relevant candidate news is appended to the basic prompt.  

This extended task, represented in dark blue in Figure ~\ref{fig:teaser}, requires the model to not only recognize but also integrate relevant context news into a cohesive narrative. Interestingly, this downstream application does more than generate a narrative by-product; it appears to refine the model's precision in the initial relevance labeling itself. For instance, in the case where the target news discusses a legal altercation between JPMorgan and Frank, a context candidate detailing JPMorgan's activities in a different region—although temporally aligned—was deemed unrelated. The \basicprompt method alone did not capture this distinction\footnote{\url{https://chat.openai.com/share/0e342312-5a14-4227-b9d6-3166b6cb5058}}, whereas the \extendedprompt approach successfully identified the lack of semantic relevance\footnote{\url{https://chat.openai.com/share/35bc4541-99be-44b2-a4ce-c6f7df6b23c4}}. 
Experiment results on existing benchmarks confirmed this observation.

While the scope of this study is confined to the domain of timeline generation from news reports, the implications of our findings extend beyond this narrow application. The observed enhancement in language model performance through the strategic use of downstream tasks presents a promising avenue for further exploration. Although it remains to be seen whether these results can be generalized across diverse tasks for large language models (LLMs), we hope that our insights into prompt engineering can inspire future research. 

%% file: sections/related_work.tex
The closest fields of our work are timeline summarization (TLS) and event extraction from news articles. Existing Natural Language Processing (NLP) studies in this area can be broadly categorized based on their core methodologies: neural network techniques for event identification and summarization, and graph-based approaches for knowledge representation and dynamic event organization.
\vspace{-2mm}
\subsection{Neural Network Techniques}
Within the domain of neural techniques, researchers have focused on developing models that are capable of capturing the nuanced relationships between events and their temporal markers. \cite{piskorski2020timelines} have pioneered the use of Abstract Meaning Representations (AMRs) to create graphical representations of text that emphasize semantic concepts and the connections between them. This approach aids in overcoming the variability of linguistic expressions, aligning sentences with different wordings but similar meanings. \cite{mansouri2023towards} have built upon this by proposing a sophisticated two-step sentence selection process that harnesses both AMRs and traditional text analysis to enhance the granularity of timeline summarization. \cite{zhou2018neural} have contributed a neural network-based framework that eschews the need for annotated datasets, which are often a bottleneck in supervised learning scenarios. Their model assumes a shared storyline distribution between article titles and bodies and across temporally adjacent documents, facilitating the autonomous generation of coherent storylines. 
\vspace{-2mm}

\subsection{Graph-Based methods}
The second approach aim to encapsulate the evolving nature of news events with a graph structure. \cite{rospocher2016building} introduce methods to automatically generate Event-Centric Knowledge Graphs (ECKGs) from news articles. These ECKGs extend beyond the static information typically found in encyclopedic knowledge graphs such as wikidata\cite{wikidata}. Story Forest\cite{liu2020story} presents a system for real-time news content organization. The system employs a semi-supervised, two-layered graph-based clustering method. StoryGraph \cite{ansah2019graph} explores the potential of graph timeline summarization by leveraging user network communities, temporal proximity, and the semantic context of events.

The field of NLP has witnessed a paradigm shift with the advent of Large Language Models (LLMs) \cite{gpt-3.5-turbo, openai2023gpt4, touvron2023llama}. These models' impressive language and reasoning capability present a new approach to the longstanding challenges. Our study diverges from established approaches by leveraging the power of LLMs to assess the relevance of news articles, suggesting a new direction for news timeline generation.

%% file: sections/architecture.tex
%

\begin{figure}[ht]
    \vspace{-5mm}

    \begin{minipage}{\linewidth}
        \centering
        \caption{system diagram}
    \includegraphics[width=\linewidth]{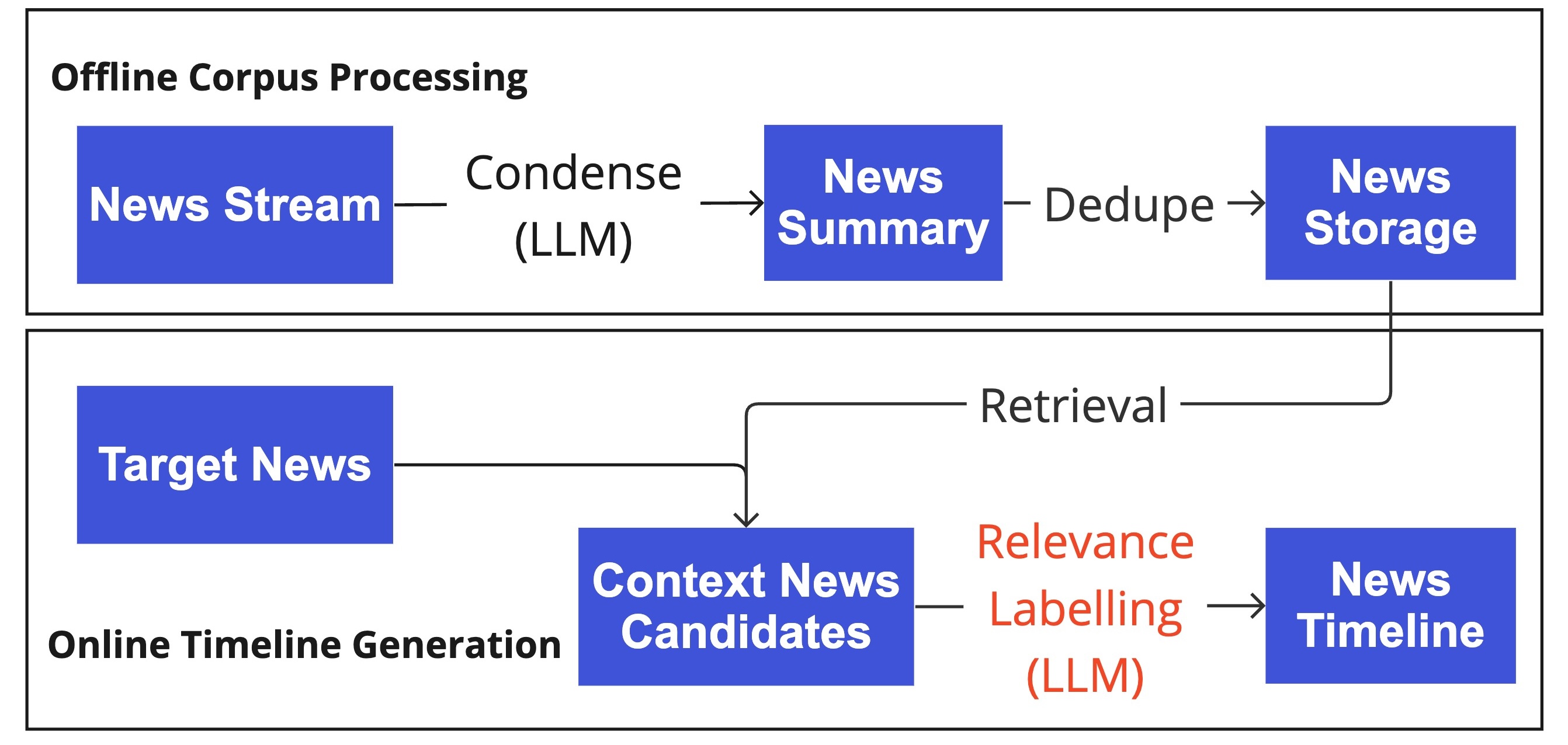}
     \vspace{-5mm}%
    \label{fig:storyline-architecture}
    \end{minipage}
   
    \begin{minipage}{\linewidth}

          \caption{case study}
    \includegraphics[width=\linewidth]{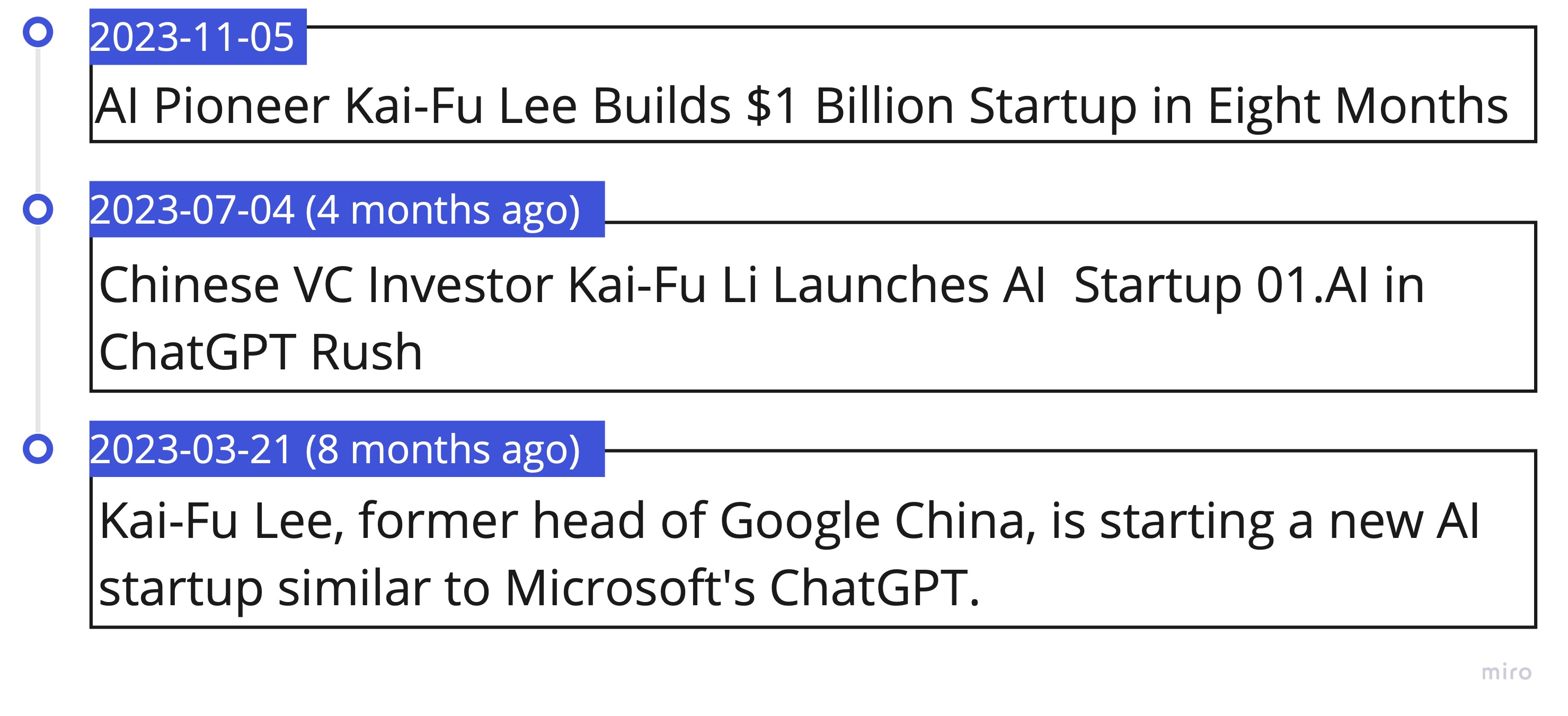}
      \vspace{-5mm}
    \label{fig:storyline-case-study}
    \end{minipage}
    \vspace{-5mm}

\end{figure}

Figure \ref{fig:storyline-architecture} outlines the dual-component system architecture designed for generating real-time news timelines. This system is segmented into an offline process for initial corpus handling and an online module that activates during user interaction with a news article via a browser plugin\footnote{\pluginsite}. \textbf{Offline Corpus Processing}:
In the offline stage, a stream of incoming news documents, denoted as $D= {d_1, d_2, ... d_t,...}$, undergoes a summarization process. Each document $d_i$ is summarized into a single sentence using a LLM, aiming to distill the core event and reduce token size. These summaries are then linked to corresponding reports from different sources. \textbf{Online Timeline Generation}: Upon a user's engagement with a target news article $d_{target}$, the online component is triggered to create a relevant timeline. It retrieves a set of context news candidates from the summarized and linked corpus stored during the offline phase. The retrieval employs a blend of existing methods \cite{mccandless2010lucene,sbert-model,yang2021newslink}, which, for the scope of this study, are treated as a black box. These context candidates are then processed alongside $d_{target}$ by the LLM, which labels each piece's relevance to the target news. Finally, the system presents a timeline $TL(d_{target})$, a chronologically arranged selection from $D$, which contextualizes $d_{target}$ within its related events. This generated timeline, as exemplified in Figure \ref{fig:storyline-case-study}, provides users with a structured historical view of the news topic at hand.

Relevance labelling is the most critical step in the whole process. Traditional retrieval methods, while adept at identifying broadly related content, often fall short in the precise curation needed within the financial sector. Financial professionals work under stringent time constraints, requiring information that is not only pertinent but also distilled to its essence. LLMs, with their advanced reasoning capabilities and contextual understanding, offer a promising solution. They can fine-tune the curation process by discerning the nuanced relationships and relevance within content, thereby automating and enhancing the accuracy of information delivery in high-stakes financial environments. 

In the quest to refine the efficacy of relevance labeling using Large Language Models (LLMs), our work has experimented with various prompt engineering techniques, notably Chain-of-Thought (CoT)\cite{wei2022chain} and Tree-of-Thought (ToT)\cite{yao2023tree}. Our exploration revealed that a step-by-step zero-shot prompting approach yielded effective results. Initially, the prompt design included only the first two steps as showcased in Figure \ref{fig:teaser}. However, we encountered instances of mislabeling, such as with the third context news candidate shown in the example.

In response to such inaccuracies, we iteratively refined our prompts. Through this process, we found that incorporating an additional step into the prompt significantly enhanced the labeling accuracy. This modification entailed requesting the LLM to generate a summary based on the entries it deemed relevant. This final step appears to have been pivotal, leading to an increase in user satisfaction with the relevance labeling task. The act of summarizing seems to encourage the LLM to more thoroughly consider the context and connections between events, resulting in a higher precision of relevance determination.

\begin{table*}[!h]

  \centering
  \include{figures/prompt_evaluation_full_one_column}
  \caption{{relevance check performance for \basicprompt and \extendedprompt}}
  \label{tab:prompt-evaluation}
  \vspace{-1cm}
\end{table*}

\begin{lstlisting}[breaklines,  caption=Storyline Prompt, label={lst:prompt-template}]
You are an experienced journalist writing a background 
story for the Target News.
  TARGET_NEWS
Here is a list of context news:
  CONTEXT_NEWS_CANDIDATES 

 Instruction:
 Step 1:Read the target news, determine the main topic, develop a short, expressive title for the story.
 
 Step 2: Select Context News. Read each piece of Context News, determine whether they are relevant to the story decided in step 1. ONLY consider news that either directly relate to or provide meaningful background to the Target News.

 Step 3 (Only in extended task prompt) Develope a short, concise, coherent background story of Target News in less than 5 sentences, using ONLY context news 'related' equals true in step 2. Provide reference to context news using the date with the format [2023-02-02].

 Formatting Your Response. Output your answer as a json object following the format below:
 {
   "storyline_title":"Short Title of the Story",
   "context_news_relevance": [{"id": 2, "related": true},..],
   "background_summary": "A concise, coherent summary providing deeper insights into the Target News, citing relevant Context News [2020-10-01] ..."    (Only in extended task prompt)
 }
\end{lstlisting}




%% file: figures/prompt_evaluation_full_one_column.tex
\begin{adjustbox}{width=0.95\linewidth}

\begin{tabular}{ |p{5cm}|ccc|ccc|ccc|ccc|} 
    \hline
     & \multicolumn{3}{c|}{Crisis} & \multicolumn{3}{c|}{TL17} & \multicolumn{3}{c|}{Finance News} & \multicolumn{3}{c|}{Overall}\\
    \hline
 & P & Recall & F1 & P & Recall & F1 & P & Recall & F1  & P & Recall & F1  
  \\
 \hline

 Vicuna-7b \basicprompt &  0.661 & 0.452 & 0.516 & 0.886 & 0.491 & 0.594 & 0.818 & 0.477 & 0.583 &  0.801 & 0.475 & 0.571
 \\
 Vicuna-7b \extendedprompt &   \textbf{0.930} & \textbf{0.593} & \textbf{0.687} & \textbf{0.945} & \textbf{0.673} & \textbf{0.738} & \textbf{0.951} & \textbf{0.620} & \textbf{0.715} & \textbf{0.945} & \textbf{0.627} & \textbf{0.714} \\
 \hline

 GPT-3.5 \basicprompt &  0.873 & 0.672 & 0.738 & 0.781 & 0.602 & 0.671 & 0.963 & 0.607 & 0.728  & 0.898 & 0.620 & 0.716 
 \\
 GPT-3.5 \extendedprompt & \textbf{0.987} & \textbf{0.711} & \textbf{0.819} & \textbf{0.962} & \textbf{0.611} & \textbf{0.730} & \textbf{0.982} & \textbf{0.637} & \textbf{0.758} & \textbf{0.978} & \textbf{0.646} & \textbf{0.764} \\

   \hline

 GPT-4 \basicprompt &  \textbf{1.000} & \textbf{0.764} & \textbf{0.865} & \textbf{1.000} & 0.739 & 0.841 & \textbf{0.993} & 0.711 & 0.817 &  \textbf{0.996} & 0.729 & 0.833 
 \\
 GPT-4 \extendedprompt & \textbf{1.000} & \textbf{0.764} & \textbf{0.865} & \textbf{1.000} & \textbf{0.748} & \textbf{0.852} & \textbf{0.993} & \textbf{0.716} & \textbf{0.827}  & \textbf{0.996} & \textbf{0.735} & \textbf{0.841} \\
 \hline
\end{tabular}
\end{adjustbox}

%% file: sections/evaluation.tex
\begin{figure}
    \vspace{-2mm}

\caption{{effects of one-shot prompt}}
    \includegraphics[width=0.55\linewidth]{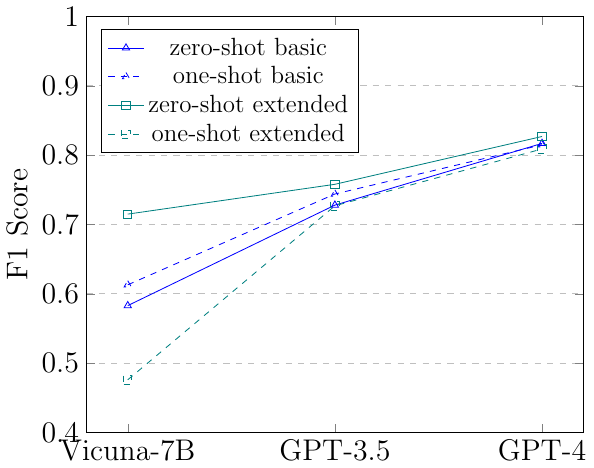}
    \label{fig:one-shot}
    \vspace{-4mm}
\end{figure}
\noindent\textbf{Deployment} We have launched a publicly accessible demonstration system available at \demosite, alongside a browser extension\footnote{\pluginsite} designed to construct real-time storylines for prominent financial websites such as FT.com, Bloomberg, Reuters, and The New York Times. Since its release in July 2023, it has become a valuable tool for our colleagues, integrating seamlessly into their workflow to enhance the consumption and understanding of financial news narratives.

\noindent\textbf{Dataset} To measure the impact of different prompt engineering strategies on LLMs for relevance labeling, we utilized established datasets such as TL17 \cite{tran2013leveraging, tran2013predicting} and crisis \cite{tran2015joint, tran2015timeline}, as well as our in-house financial news collection. The TL17 and crisis datasets, relevance is labelled by human. For the financial dataset, relevance was deduced from internal hyperlinks within articles. Given the LLMs' limitations on context size, we selected five articles from each timeline as positive samples. For negative samples, we chose articles from similar periods but ensured a clear semantic distinction, indicated by a cosine similarity lower than 0.1 of embeddings calculated by sbert\cite{sbert-model}. This approach yielded a total of 88 timelines: 22 from TL17, 19 from crisis, and 47 from financial news—offering a broad spectrum for our LLM relevance labeling evaluation. This dataset can be downloaded at \footnote{\url{https://www.notion.so/News-Storyline-952185b5a997461c9750ab3fbb202a75}}.

\noindent\textbf{Large Language Models} In our experiments, we have employed three different LLMs: Vicuna-7b-v1.5\cite{zheng2023judging}, GPT-3.5-turbo\cite{gpt-3.5-turbo}, and GPT-4\cite{openai2023gpt4}. Vicuna-7b is an open-source model derived by fine-tuning from Llama 2\cite{touvron2023llama}, with a capacity of 7 billion parameters. It's the smallest model we tried so far that can consistently output response in required format for automation tasks. GPT-3.5-turbo is recognized for its efficiency, providing a balance of performance and affordability for a wide array of linguistic tasks. The most advanced among them, GPT-4, is at the forefront of current LLM technology, offering state-of-the-art capabilities. We accessed Vicuna-7b through the Hugging Face platform\footnote{\url{https://huggingface.co/lmsys/vicuna-7b-v1.5}} and made use of the official APIs provided by OpenAI for GPT-3.5 and GPT-4.

\noindent\textbf{Prompt Messages} Both positive and negative entries are mixed together and sorted chronologically before feeding into prompt template. All news timestamp are also included in the prompt. The output is a json object. The prompt templates can be found in Listing \ref{lst:prompt-template}. \basicprompt only contains Step 1 and Step 2. \extendedprompt has an additional Step 3.

\noindent\textbf{Result}
Table \ref{tab:prompt-evaluation} shows the comparative efficacy of two prompt templates across varied content domains: crisis events, TL17, and financial news. The \extendedprompt demonstrates superior F1 scores across all language models for each dataset examined. While the \basicprompt result in high precision, they are deficient in recall. This indicates that although the predictions are precise, they likely miss many relevant articles. In contrast, the \extendedprompt exhibit a more robust performance profile, with elevated precision and recall that culminate in higher F1 scores across all models. This trend suggests that engaging LLMs with summary generation prompts may facilitate a more exhaustive evaluation of article relevance, leading to a more equitable selection of news articles. The performance boost conferred by the \extendedprompt is notably more significant for the less advanced models than advanced models. 
Figure \ref{fig:one-shot} delves into the influence of one-shot in-context learning on model performance, with the dotted lines charting the F1 scores for one-shot prompts. It is observed that the \basicprompt maintains a similar performance in both one-shot and zero-shot setups. However, the \extendedprompt is adversely affected by one-shot prompting, a phenomenon more pronounced in less capable models. A detailed examination reveals that Vicuna-7b becomes more cautious in issuing \quoted{related} labels post-exposure to the example. This could be attributed to the fact that the one-shot example contains only two related articles, whereas the experimental data averages five positive cases.

%% file: sections/conclusion.tex
In our study, we have presented a prompt-engineering technique that significantly enhances the process of generating news storylines. By employing an \extendedprompt, we have enabled large language models (LLMs) to discern subtle semantic variations within news content, which has substantially increased the uptake of news timeline applications among financial professionals. We hope this research will not only garner interest but also stimulate further exploration in the realms of event timeline construction and the refinement of LLM prompting strategies.